\documentclass[11pt]{article}

\PassOptionsToPackage{table}{xcolor}
\usepackage[preprint]{acl}
\usepackage{subcaption}
\usepackage{times}
\usepackage{latexsym}
\usepackage[T1]{fontenc}
\usepackage[utf8]{inputenc}
\usepackage{microtype}
\usepackage{inconsolata}
\usepackage{graphicx}
\usepackage{float}

\usepackage{adjustbox}

\usepackage{amsmath}
\usepackage{amssymb}
\usepackage{mathtools}
\usepackage{bm}

\usepackage{booktabs}
\usepackage{multirow}
\usepackage{makecell}
\usepackage{array}
\usepackage{tabularx}

\usepackage{pifont}            
\usepackage[normalem]{ulem} 

\usepackage{algorithm}
\usepackage{algpseudocode}

\usepackage{enumitem}
\usepackage{xspace}

\title{{\fontsize{16pt}{19pt}\selectfont BOSS: Token-Budgeted Online Sampling\\and Selection for LLM Reasoning}}

\author{
{\fontsize{12pt}{14.5pt}\selectfont\rmfamily\bfseries
Yupeng Chang$^{1}$ \quad Yahan Li$^{4}$ \quad Chenlu Guo$^{5}$} \\
{\fontsize{12pt}{14.5pt}\selectfont\rmfamily\bfseries
Diana Inkpen$^{6}$ \quad Yuan Wu$^{1}$\footnotemark[1] \quad Yi Chang$^{1,2,3}$} \\[0.38em]
{\fontsize{11pt}{13.6pt}\selectfont\rmfamily\mdseries
$^{1}$School of Artificial Intelligence, Jilin University} \\
{\fontsize{11pt}{13.6pt}\selectfont\rmfamily\mdseries
$^{2}$Engineering Research Center of Knowledge-Driven Human-Machine Intelligence, MOE, China} \\
{\fontsize{11pt}{13.6pt}\selectfont\rmfamily\mdseries
$^{3}$International Center of Future Science, Jilin University} \\
{\fontsize{11pt}{13.6pt}\selectfont\rmfamily\mdseries
$^{4}$Xingchen AGI Lab, China Telecom Artificial Intelligence Technology (Beijing) Co., Ltd., China} \\
{\fontsize{11pt}{13.6pt}\selectfont\rmfamily\mdseries
$^{5}$Bilibili Inc., China} \\
{\fontsize{11pt}{13.6pt}\selectfont\rmfamily\mdseries
$^{6}$School of Electrical Engineering and Computer Science, University of Ottawa, Canada} \\[0.25em]
{\fontsize{11pt}{13.6pt}\selectfont\rmfamily\mdseries
changyp23@mails.jlu.edu.cn, \{yuanwu, yichang\}@jlu.edu.cn} \\
{\fontsize{11pt}{13.6pt}\selectfont\rmfamily\mdseries
liyh181@chinatelecom.cn, guochenlu01@bilibili.com, Diana.Inkpen@uottawa.ca}
}

\newcommand{\BOSSOpenReviewPaperID}{}
\newif\ifBOSSOpenReviewPublic
\BOSSOpenReviewPublicfalse  
\newcommand{\BOSSOpenReviewLine}{%
  \ifBOSSOpenReviewPublic
    \space OpenReview: \url{https://openreview.net/forum?id=\BOSSOpenReviewPaperID}.
  \fi
}

\hypersetup{
  pdftitle={BOSS: Token-Budgeted Online Sampling and Selection for LLM Reasoning},
  pdfauthor={Yupeng Chang, Yahan Li, Chenlu Guo, Diana Inkpen, Yuan Wu, Yi Chang},
  pdfsubject={arXiv preprint; test-time reasoning and inference token budgets}
}
\begin{document}
\maketitle
\renewcommand{\thefootnote}{\fnsymbol{footnote}}
\footnotetext[1]{Corresponding author}
\renewcommand{\thefootnote}{\arabic{footnote}}

\begin{abstract}
Test-time scaling can improve LLM reasoning, but existing approaches typically adapt generation or apply answer selection after a predetermined generation schedule. We study \emph{within-instance allocation}: under a hard per-instance prompt-plus-completion token cap, when should inference stop, generate another candidate, or spend residual tokens adjudicating candidates already observed? BOSS is a training-free policy for this three-way decision, with worst-case budget checks before every generation and judge call. Its key contribution is online generation--adjudication allocation under one shared token ledger. Across two Llama-3.1 models and four reasoning benchmarks, BOSS attains the highest accuracy in 22 of 24 model--task--budget cells under the common hard cap. At Llama-3.1-8B and $B=4\mathrm{k}$, it reaches 60.5 macro-average accuracy versus 58.6 for Adaptive-Consistency, 59.1 for TALE-EP, and 58.9 for fixed-schedule League. On identical Stage-I candidate pools, randomized A/B judging with repeated voting improves plurality from 58.4 to 60.3, with the full BOSS selector reaching 60.5. Approximately matched-cost controls further characterize how the benefit of adjudication changes with available inference budget. Together, these results establish generation--adjudication allocation as a distinct design axis for reasoning under strict per-instance caps. Code: \url{https://anonymous.4open.science/r/boss-BF00/}.\BOSSOpenReviewLine
\end{abstract}

\section{Introduction}
Test-time scaling (TTS) improves LLM reasoning by spending additional inference compute, but the same nominal budget can be used in qualitatively different ways: to generate more candidate solutions, to lengthen or parallelize reasoning, or to compare answers that have already been generated~\citep{wang2022selfconsistency,snell2024scalingttc,zhang2025surveytts}. This distinction matters when deployment imposes a hard per-instance token cap. A token spent on another reasoning trace cannot also be spent adjudicating among existing answer hypotheses, and comparison calls themselves incur prompt cost because they must include the problem and candidate evidence.

Prior work addresses important parts of this design space, but not the allocation question studied here. Adaptive-Consistency and ESC decide \emph{whether further sampling is necessary} and then aggregate the sampled answers~\citep{aggarwal2023adaptive,li2024esc}. TALE-EP and budget-forcing methods adapt \emph{trajectory length or token allowance}~\citep{han2025tale,muennighoff2025s1}. Difficulty-adaptive self-consistency allocates sampling effort across questions, while Parallel-Probe controls width and depth in parallel reasoning~\citep{wang2025dsc,zheng2026parallelprobe}. Pairwise selection methods, in contrast, spend test-time compute comparing candidates, but typically do so after a fixed or separately tuned generation schedule~\citep{chen2026provable}. These are complementary choices. We focus on the within-instance decision: \textbf{given a hard per-instance cap, how should one instance allocate its realized token budget online between additional candidate generation and answer adjudication?}

We formulate this as a three-way online decision. After each generated candidate, an inference policy may (i) return a sufficiently stable answer, (ii) generate again because useful hypotheses are still emerging, or (iii) redirect the residual budget to adjudicating hypotheses already in the candidate pool. BOSS implements this decision with consensus-based termination, a novelty-based switch criterion, and budget-feasible pairwise judging. All prompt and completion tokens---including the full judge prompt containing the original problem and both representatives---are charged to the same ledger. Before every call, BOSS checks the prompt length plus the maximum completion cap, so adaptive decisions cannot overspend the per-instance budget.

BOSS builds on established stopping, grouping, and judging primitives, but couples them inside a single \emph{online generation--adjudication allocator}, where selection is an alternative use of the residual budget rather than a fixed post-processing stage. This allocation decision is the methodological contribution. Its effect is directly testable: a development-tuned fixed generation--selection split reaches 59.3 macro-average at Llama-3.1-8B, $B=4\mathrm{k}$, versus 60.5 for BOSS, and disabling the novelty-triggered switch reaches 59.0.

Our experiments are designed to isolate the value of online allocation. First, we compare against Adaptive-Consistency and TALE-EP under the same strict token ledger. Second, we reuse exactly the same Stage-I candidate pools and change only the selector, separating candidate-generation quality from the value of adjudication. Third, we complement the hard-cap evaluation with realized token consumption and approximately cost-matched operating points. Together, these controls distinguish the effect of the allocator from adaptive stopping, candidate quality, and differences in realized token use. The strongest controlled result is the identical-pool comparison: without changing any Stage-I candidates, randomized repeated judging raises macro accuracy from 58.4 to 60.3, and the full BOSS selector reaches 60.5.

We summarize the contributions as follows:
\begin{itemize}
    \item \textbf{A budget-faithful within-instance allocation problem.} We formalize generation and adjudication as competing actions under one realized prompt-plus-completion ledger and enforce worst-case feasibility before every model call.
    \item \textbf{A three-way online policy.} BOSS chooses among returning, continuing candidate generation, and entering answer adjudication based on observed consensus, answer novelty, comparison feasibility, and residual budget.
    \item \textbf{Evidence isolating the allocator's contribution.} Beyond hard-cap results on two model scales and four benchmarks, we include adaptive-stopping and token-budget baselines, same-candidate-pool selector controls, realized-cost comparisons, transition statistics, hyperparameter sensitivity, and paired per-instance tests. These studies show that the gains arise from adaptive allocation and adjudication rather than a stronger candidate pool or uncounted selection overhead, while characterizing the budget regimes in which adjudication is most useful.
\end{itemize}

\section{Related Work}
\paragraph{Adaptive generation.} Self-consistency scales inference by sampling multiple traces~\citep{wang2022selfconsistency}. Adaptive-Consistency, ESC, and Stop When Enough study adaptive stopping of sampled or chain-of-thought reasoning~\citep{aggarwal2023adaptive,li2024esc,sun2510stop}, while Difficulty-Adaptive Self-Consistency allocates sampling effort across queries using difficulty~\citep{wang2025dsc}. TALE-EP adapts reasoning-token budgets and s1 uses budget forcing to control trajectory length~\citep{han2025tale,muennighoff2025s1}; Parallel-Probe dynamically regulates width and depth of parallel thinking~\citep{zheng2026parallelprobe}. These methods primarily control generation effort. BOSS instead fixes the per-instance cap and asks when residual tokens should stop generating and begin adjudicating existing answer hypotheses.

\paragraph{Candidate selection.} Selection and verification methods spend inference compute comparing candidates~\citep{chen2026provable,chen2026rethinking,lin2026caps}. LLM judges are flexible but position-sensitive, motivating randomized order and repeated judgments~\citep{zheng2023mtbench,shi2024judging}. BOSS uses these established selection primitives differently: selection becomes an online alternative to further generation, with its start and extent determined by the realized answer distribution, novelty trajectory, representative-prompt cost, and residual budget. We test this distinction with a fixed-split ablation and selectors applied to identical Stage-I candidate pools.

\paragraph{Scope.} Cross-instance allocation and within-instance generation--adjudication allocation are complementary. BOSS targets the latter: online allocation between additional generation and answer adjudication under a strict per-instance cap.

\section{Method}
\label{sec:method}

\begin{figure*}[t]
  \centering
  \includegraphics[width=0.85\textwidth]{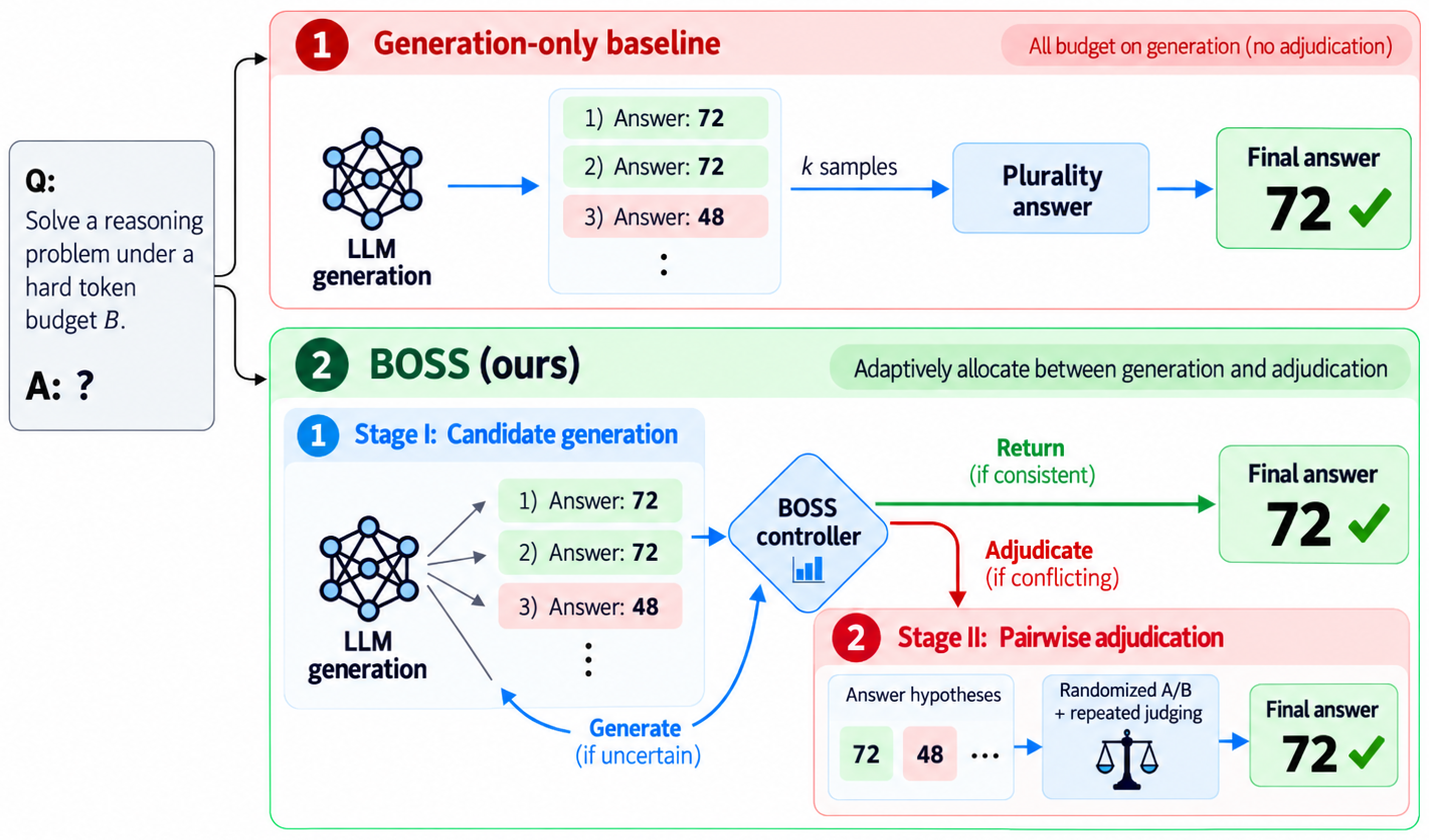}
  \caption{\textbf{Overview of BOSS.} Under the same hard cap $B$, BOSS allocates residual tokens online among \textsc{Return}, \textsc{Generate}, and \textsc{Adjudicate}.}
  \label{fig:boss_overview}
\end{figure*}

\subsection{Problem Setup and Budget Safety}

We study test-time reasoning under a hard per-instance token budget. Given an input instance $x$, an LLM produces a completion
\begin{equation}
y \sim p_\theta(\cdot \mid p),
\end{equation}
where $p$ is the prompt. A deterministic parser extracts the final answer
\begin{equation}
\phi:\mathcal{V}^{*}\rightarrow \mathcal{A}\cup\{\bot\},
\end{equation}
where $\bot$ denotes parsing failure and is counted as incorrect. We focus on reasoning tasks with deterministically parsable final answers, such as exact-match mathematical reasoning and multiple-choice reasoning.

Inference consists of an adaptive sequence of LLM calls. Call $t$ uses prompt $p_t$, returns completion $y_t$, and consumes
\begin{equation}
c_t=\mathrm{tok}(p_t)+\mathrm{tok}(y_t),
\end{equation}
where $\mathrm{tok}(\cdot)$ is computed with the deployed tokenizer. Under a hard token budget $B$, every realized inference trajectory must satisfy
\begin{equation}
\sum_{t=1}^{T} c_t \le B.
\end{equation}
Thus, generation and selection calls must share the same prompt-plus-completion token ledger.

Each call is executed with a maximum new-token cap. Generation calls use cap $L^{\mathrm{gen}}$, and judge calls use cap $L^{\mathrm{judge}}$. For a prompt $p$ and cap $L$, define the worst-case cost
\begin{equation}
\overline{c}(p;L)=\mathrm{tok}(p)+L.
\end{equation}
A call is issued only if the current cumulative cost $C$ satisfies
\begin{equation}
C+\overline{c}(p;L)\le B.
\end{equation}
Because the actual completion length never exceeds the cap, this check guarantees that realized token use does not exceed $B$.

\paragraph{Allocation state and action space.}
The budget constraint alone does not specify how the remaining tokens should be used. After observing a candidate pool $S_t$ and cumulative realized cost $C_t$, we view the policy decision as
\begin{equation}
u_t\in\{\textsc{Return},\textsc{Generate},\textsc{Adjudicate}\}.
\end{equation}
\textsc{Return} terminates with the current answer, \textsc{Generate} spends budget on another candidate, and \textsc{Adjudicate} freezes the current candidate pool and spends the residual budget on answer-level comparisons. BOSS is one concrete, training-free policy for choosing among these actions. This formulation is useful because it isolates the paper's allocation question from the specific stopping and judging primitives used to implement it.

\subsection{BOSS: Budgeted Online Sampling and Selection}

We propose \textbf{BOSS} (\textbf{B}udgeted \textbf{O}nline \textbf{S}ampling and \textbf{S}election), a training-free inference policy that adaptively allocates a shared token budget between candidate sampling and answer adjudication. BOSS samples while useful hypotheses are still emerging, returns when a stable consensus forms, and shifts residual budget to randomized-order repeated pairwise judging once recent samples introduce little new answer diversity and adjudication is budget-feasible.

BOSS uses two black-box LLM primitives. A \textsc{Generate} call samples a candidate completion and parses its answer; a \textsc{Judge} call compares representative completions for two answer hypotheses. By default both use the same base model, so improvements come from test-time allocation rather than an external verifier. The complete step-by-step procedure appears as Algorithm~\ref{alg:boss} in Appendix~\ref{sec:detailed_scheduling}.

The policy has three actions. \textsc{Return} is appropriate when additional sampling is unlikely to change a stable answer; \textsc{Generate} is useful while new answer hypotheses are still appearing; and \textsc{Adjudicate} becomes attractive when recent answer novelty is low, multiple plausible hypotheses remain, and the residual budget can support reliable comparisons. This action-level view is the central design choice of BOSS: the method does not reserve a fixed fraction of the budget for selection in advance.

\paragraph{Why consensus and novelty are separate signals.}
A high consensus ratio and low recent novelty are not interchangeable. Strong consensus supports \textsc{Return}: the candidate pool already points to one answer, so further sampling or judging is less necessary. Low novelty without consensus describes a different state: repeated sampling is revisiting existing hypotheses, yet no answer is sufficiently dominant. In that regime, spending more tokens on additional traces that map to already observed answer hypotheses can be less useful than comparing the competing hypotheses already available. BOSS therefore uses consensus to decide when the candidate pool is sufficiently stable, and novelty to decide when additional generation is yielding fewer new hypotheses.

\paragraph{Why the switch is gated by feasibility.}
The switch criterion is coupled to the token ledger rather than evaluated in isolation. BOSS enters adjudication only if the remaining budget can pay for a complete comparison prompt and its capped output; otherwise the selection stage would be infeasible. Conversely, BOSS does not switch merely because two answers exist. When novelty remains high, generation can still expand the hypothesis set, and the policy continues sampling if the next call is feasible. These two gates---evidence that generation is saturating and evidence that comparison is affordable---make the transition budget-aware rather than a fixed stage boundary. They also explain why the method can behave differently across budget regimes without changing hyperparameters.

\subsection{Stage I: Adaptive Candidate Sampling}

BOSS first samples candidate solutions
\begin{equation}
S_n=\{(y_i,a_i)\}_{i=1}^{n},\qquad a_i=\phi(y_i).
\end{equation}
For each parsed answer $a\in\mathcal{A}$, let $n_a$ be its frequency among valid parsed candidates, and let
\begin{equation}
n_{\mathrm{parsed}}=\sum_{a\in\mathcal{A}}n_a.
\end{equation}
Denote the plurality answer by $a^{(1)}$ and the runner-up count by $n^{(2)}$, with $n^{(2)}=0$ when fewer than two distinct parsed answers are present.

When $n_{\mathrm{parsed}}>0$, BOSS returns early when one answer becomes sufficiently dominant:
\begin{equation}
\frac{n_{a^{(1)}}}{n_{\mathrm{parsed}}}\ge \tau_{\mathrm{cons}}
\quad\text{and}\quad
n_{a^{(1)}}-n^{(2)}\ge m_{\mathrm{gap}}.
\end{equation}
This consensus rule avoids spending additional tokens when repeated sampling already supports a stable answer. If $n_{\mathrm{parsed}}=0$, the consensus condition is false.

If no consensus has formed, BOSS checks whether recent samples are still producing new answer hypotheses. Let
\begin{equation}
\mathcal{U}_{\le t}=\{a_i\in\mathcal{A}: i\le t,\ a_i\neq\bot\}
\end{equation}
be the set of distinct parsed answers observed up to step $t$. We evaluate the novelty rule after at least $W$ samples, with $\mathcal{U}_{\le 0}=\emptyset$ for boundary cases. With window size $W$, define
\begin{equation}
\mathrm{novel}(t)=
\frac{\left|\mathcal{U}_{\le t}\setminus\mathcal{U}_{\le(t-W)}\right|}{W}.
\end{equation}
BOSS transitions to Stage~II when
\begin{equation}
\mathrm{novel}(t)\le \rho_{\mathrm{novel}}
\quad\text{and}\quad
|\mathcal{U}_{\le t}|\ge 2,
\end{equation}
provided that at least one complete pairwise comparison is budget-feasible. The rule uses diminishing answer novelty as a lightweight empirical signal that the marginal value of another sample may be lower than the value of adjudicating existing alternatives. Section~\ref{sec:robustness_transition} reports transition rates and sensitivity to the switching hyperparameters.

If the novelty-triggered switch has not fired and the next generation call is infeasible, BOSS returns the plurality parsed answer, or $\bot$ if no valid parsed answer exists. Thus, budget exhaustion followed by plurality is a safety fallback rather than an implicit trigger for Stage~II.

\subsection{Stage II: Budgeted Pairwise Selection}

Stage~II operates over parsed answer hypotheses rather than all individual completions. Let
\begin{equation}
\mathcal{H}_x=\{a\in\mathcal{A}:n_a>0\}.
\end{equation}
To control comparison cost, BOSS keeps the $K_{\max}$ most frequent hypotheses:
\begin{equation}
\mathcal{H}\leftarrow \mathrm{TopK}(\mathcal{H}_x;K_{\max}).
\end{equation}
For each retained answer $a\in\mathcal{H}$, BOSS selects a compact representative completion $\mathrm{rep}(a)$ from the candidates supporting that answer. Long representatives are truncated while preserving the final-answer line, reducing comparison-prompt overhead.

BOSS spends the residual budget on pairwise comparisons. To compare two hypotheses $a$ and $b$, it presents their representatives to the judge with randomized A/B ordering and repeats the comparison up to $R$ times when budget permits. The winner is determined by majority vote over valid judge outputs, with abstentions ignored. Randomized ordering and repeated voting reduce sensitivity to presentation-order bias in LLM-as-a-judge comparisons~\citep{zheng2023mtbench,shi2024judging}.

Because the number of affordable comparisons varies by instance, BOSS schedules comparisons online. It maintains a smoothed win estimate for each active hypothesis,
\begin{equation}
\hat{p}_a=\frac{w_a+\alpha}{g_a+2\alpha},
\end{equation}
where $w_a$ is the number of wins assigned to answer $a$, $g_a$ is the number of matches involving $a$, and $\alpha$ is a smoothing constant. BOSS prioritizes comparisons among hypotheses that remain plausible under the current match outcomes, using lightweight uncertainty-guided scheduling inspired by noisy pairwise preference feedback~\citep{yue2012duelingbandits,zoghi2014rucb}. This scheduling rule is used only to allocate the residual budget efficiently, not as a calibrated statistical confidence bound.

When no further worst-case-feasible judge call can be issued, selection stops. BOSS returns the active hypothesis with the highest $\hat{p}_a$. Remaining ties are broken deterministically by larger Stage-I frequency, shorter representative length, and then lexicographic order. If no pairwise comparison is executed, BOSS falls back to deterministic plurality selection over the sampled candidates.

\section{Experiments}
\label{sec:experiments}

\subsection{Experimental Setup}

\paragraph{Models.}
We evaluate BOSS on two open-weight instruction-tuned LLMs at two model scales: \emph{Llama-3.1-8B-Instruct} and \emph{Llama-3.1-70B-Instruct}~\citep{grattafiori2024llama3,llama31modelcard8b,llama31modelcard70b}. Unless otherwise noted, \textsc{Generate} and \textsc{Judge} use the same base model. Thus, performance differences reflect test-time allocation rather than the use of a stronger external verifier.

\paragraph{Hard-cap protocol and accounting.}
All methods operate under the same hard per-instance token cap $B$. We count \emph{every} prompt and completion token from generation and selection. In particular, each judge prompt contains the original problem plus both representative completions, and the full prompt cost is charged on every replicate. A call is issued only if the remaining budget covers the full prompt plus its maximum completion cap. For parallel methods such as FFS and short-$m@k$, tokens from all active streams are charged through termination, including unfinished streams that do not provide the returned answer. The main hard-cap results use $B\in\{2\mathrm{k},4\mathrm{k},8\mathrm{k}\}$. Because equal caps do not imply equal realized cost, Section~\ref{sec:realized_cost} separately reports mean realized tokens and approximately cost-matched operating points.

\paragraph{Decoding and judging.}
\textsc{Generate} uses nucleus sampling with temperature $\tau=0.7$ and top-$p=0.95$, capped at $L^{\mathrm{gen}}=1024$ new tokens. \textsc{Judge} uses deterministic decoding with temperature $0$, capped at $L^{\mathrm{judge}}=4$, and constrained decoding to $\{\texttt{A},\texttt{B}\}$. Invalid outputs are treated as abstentions. Unless otherwise stated, each pairwise match uses randomized A/B ordering and majority vote over up to $R=3$ judge replicates, subject to the remaining token budget. This judging interface is shared by BOSS and all comparison-based baselines.

\paragraph{Development tuning, seeds, and paired analysis.}
All allocation-sensitive hyperparameters are selected on held-out development subsets disjoint from the reported evaluation data and are then fixed before test evaluation. This protocol is applied uniformly to BOSS and tunable baselines, including the parallelism parameters of FFS and short-$m@k$, fixed generation--selection splits, and BOSS transition/selection parameters. All stochastic methods use 5 random seeds. Tables report mean accuracy unless otherwise stated. For the closest Llama-3.1-8B comparisons at $B=4\mathrm{k}$, we additionally use paired per-instance tests against the strongest non-BOSS baseline for each benchmark.

\subsection{Benchmarks and Metrics}

We evaluate on four reasoning benchmarks with deterministically parsable final answers: \textbf{GSM8K}~\citep{cobbe2021gsm8k}, \textbf{MATH}~\citep{hendrycks2021math}, \textbf{GPQA Diamond}~\citep{rein2023gpqa}, and \textbf{MMLU-Pro}~\citep{wang2024mmlupro}. These benchmarks cover grade-school mathematical reasoning, competition-level mathematics, graduate-level scientific reasoning, and multi-domain reasoning.

For GSM8K and MATH, we report exact match (EM) after standard answer normalization, including removal of units when applicable, commas, formatting wrappers, and common \LaTeX{} markup artifacts. For GPQA Diamond and MMLU-Pro, we report multiple-choice accuracy. In all cases, the final prediction is obtained through the same task-specific parser $\phi$, and parsing failures $\bot$ are counted as incorrect. The same parsing and normalization procedures are applied to BOSS and all baselines.

\subsection{Baselines}
We implement all baselines with the same token ledger, decoding caps, parsing interface, and feasibility checks. \textbf{CoT@1} uses one deterministic generation. \textbf{Self-Consistency (SC)} samples until another call is infeasible and then uses majority voting~\citep{wang2022selfconsistency}; \textbf{ESC}~\citep{li2024esc} and \textbf{Adaptive-Consistency}~\citep{aggarwal2023adaptive} adapt when sampling stops but do not redirect residual compute to pairwise adjudication. \textbf{TALE-EP} adapts reasoning length to problem complexity~\citep{han2025tale}. \textbf{FFS} and \textbf{short-$m@k$} use parallel generation while charging every active stream~\citep{agarwal2025ffs,hassid2025dontoverthink}. \textbf{Knockout} and \textbf{League} are fixed-schedule pairwise-selection baselines~\citep{chen2026provable}; they use the same representatives, constrained judge, A/B randomization, repeated voting, and accounting as BOSS. We report Adaptive-Consistency and TALE-EP in the targeted $B=4\mathrm{k}$ analysis under the identical Llama-3.1-8B accounting protocol.

\begin{table*}[t]
\caption{Accuracy (\%) under fixed per-instance token budgets $B$. The budget counts all prompt and completion tokens consumed by each inference policy. Values are averaged over 5 random seeds. CoT@1 (Greedy) is budget-invariant because one deterministic generation uses far fewer tokens than the smallest budget ($2\mathrm{k}$); identical values across budgets are shown in merged cells. GPQA-D denotes GPQA-Diamond. Bold indicates the best result within each model block and column.}
\label{tab:main_results}
\centering
\scriptsize
\resizebox{\textwidth}{!}{%
\begin{tabular}{@{}lcccccccccccc@{}}
\toprule
& \multicolumn{3}{c}{\textbf{GSM8K} (EM)}
& \multicolumn{3}{c}{\textbf{MATH} (EM)}
& \multicolumn{3}{c}{\textbf{GPQA-D} (Acc)}
& \multicolumn{3}{c}{\textbf{MMLU-Pro} (Acc)} \\
\cmidrule(lr){2-4}\cmidrule(lr){5-7}\cmidrule(lr){8-10}\cmidrule(lr){11-13}
Method
& $2\mathrm{k}$ & $4\mathrm{k}$ & $8\mathrm{k}$
& $2\mathrm{k}$ & $4\mathrm{k}$ & $8\mathrm{k}$
& $2\mathrm{k}$ & $4\mathrm{k}$ & $8\mathrm{k}$
& $2\mathrm{k}$ & $4\mathrm{k}$ & $8\mathrm{k}$ \\
\midrule
\multicolumn{13}{l}{\textit{\textbf{Base model: Llama-3.1-8B-Instruct}}} \\
\midrule
CoT@1 (Greedy)
& \multicolumn{3}{c}{84.5}
& \multicolumn{3}{c}{51.9}
& \multicolumn{3}{c}{30.4}
& \multicolumn{3}{c}{48.3} \\

SC (budgeted)
& 85.8 & 87.3 & 88.1
& 54.7 & 56.6 & 58.0
& 33.1 & 35.7 & 37.0
& 49.9 & 51.2 & 52.2 \\

ESC \citep{li2024esc}
& 86.2 & 87.6 & 88.3
& 55.3 & 56.8 & 58.4
& 33.5 & 36.1 & 37.2
& 50.3 & 51.6 & 52.5 \\

FFS \citep{agarwal2025ffs}
& \textbf{86.9} & 87.2 & 87.5
& 54.5 & 55.7 & 56.0
& 32.2 & 33.3 & 33.8
& 49.5 & 50.1 & 50.8 \\

short-$m@k$ \citep{hassid2025dontoverthink}
& 86.5 & 87.9 & 88.5
& 55.0 & 57.1 & 58.7
& 33.4 & 35.9 & 37.5
& 50.1 & 51.8 & 53.0 \\

Knockout \citep{chen2026provable}
& 84.3 & 87.1 & 88.7
& 51.8 & 56.4 & 59.2
& 31.2 & 37.0 & 40.1
& 48.8 & 52.3 & 54.4 \\

League \citep{chen2026provable}
& 84.8 & 87.5 & 89.2
& 52.6 & 57.2 & 59.8
& 31.7 & 37.8 & 40.8
& 49.4 & 52.9 & 55.1 \\

\rowcolor{gray!10}
\textbf{BOSS (Ours)}
& 86.6 & \textbf{88.8} & \textbf{89.9}
& \textbf{56.1} & \textbf{59.4} & \textbf{61.3}
& \textbf{34.4} & \textbf{39.5} & \textbf{42.3}
& \textbf{51.2} & \textbf{54.4} & \textbf{56.3} \\

\midrule
\multicolumn{13}{l}{\textit{\textbf{Base model: Llama-3.1-70B-Instruct}}} \\
\midrule
CoT@1 (Greedy)
& \multicolumn{3}{c}{95.1}
& \multicolumn{3}{c}{68.0}
& \multicolumn{3}{c}{46.7}
& \multicolumn{3}{c}{66.4} \\

SC (budgeted)
& 95.9 & 96.5 & 96.9
& 71.1 & 72.8 & 73.6
& 48.8 & 50.4 & 51.3
& 68.3 & 69.8 & 70.6 \\

ESC \citep{li2024esc}
& 96.1 & 96.8 & 97.1
& 71.4 & 73.1 & 73.9
& 49.2 & 50.7 & 51.6
& 68.5 & 70.0 & 70.9 \\

FFS \citep{agarwal2025ffs}
& \textbf{96.5} & 96.6 & 96.7
& 70.6 & 71.7 & 72.2
& 48.3 & 49.1 & 49.6
& 67.8 & 68.5 & 69.0 \\

short-$m@k$ \citep{hassid2025dontoverthink}
& 96.2 & 96.9 & 97.3
& 71.7 & 73.4 & 74.6
& 49.1 & 51.0 & 51.9
& 68.8 & 70.2 & 71.3 \\

Knockout \citep{chen2026provable}
& 95.0 & 96.4 & 97.2
& 69.1 & 73.6 & 75.4
& 47.8 & 51.5 & 53.1
& 67.6 & 71.0 & 72.6 \\

League \citep{chen2026provable}
& 95.2 & 96.7 & 97.5
& 69.8 & 74.3 & 76.1
& 48.5 & 52.1 & 53.7
& 68.1 & 71.6 & 73.2 \\

\rowcolor{gray!10}
\textbf{BOSS (Ours)}
& 96.3 & \textbf{97.2} & \textbf{97.7}
& \textbf{72.3} & \textbf{75.5} & \textbf{77.6}
& \textbf{50.4} & \textbf{52.9} & \textbf{54.6}
& \textbf{69.5} & \textbf{72.1} & \textbf{73.8} \\

\bottomrule
\end{tabular}%
}
\end{table*}

\subsection{Main Results under Hard Token Caps}
\label{sec:main_results_analysis}

Table~\ref{tab:main_results} reports accuracy under the same hard per-instance cap. This is a \emph{cap-matched} comparison: every method receives the same deployment allowance. BOSS obtains the highest accuracy in 22 of 24 model--task--budget settings. At $B\in\{4\mathrm{k},8\mathrm{k}\}$, it is best in every column across both model scales and all four benchmarks; at $2\mathrm{k}$, it leads 6 of 8 model--task cells, with FFS leading the two GSM8K cells. The pattern persists across model scale, and the clearest margins occur on MATH and GPQA-Diamond, where competing answer hypotheses leave more room for adjudication than on the more saturated GSM8K benchmark.

For Llama-3.1-8B at $B=4\mathrm{k}$, paired per-instance tests against the strongest non-BOSS baseline on each benchmark yield improvements of $+0.9$ points on GSM8K ($p=0.021$), $+2.2$ on MATH ($p<0.001$), $+1.7$ on GPQA-Diamond ($p=0.026$), and $+1.5$ on MMLU-Pro ($p<0.001$). All four paired comparisons are significant at $p<0.05$; Appendix~\ref{sec:additional_diagnostics} reports the corresponding accuracy intervals.

\subsection{Targeted Controls for Allocation and Selection}
\label{sec:targeted_controls}

To separate the sources of the hard-cap gains, Table~\ref{tab:targeted_controls} combines complementary controls at Llama-3.1-8B, $B=4\mathrm{k}$. Adaptive-Consistency and TALE-EP test alternative adaptive uses of inference tokens, while the fixed-split and no-novelty variants test the online transition policy. The final block holds the entire Stage-I candidate pool fixed and changes only the selector, directly separating candidate quality from selection quality.

\begin{table}[t]
\centering
\small
\caption{Targeted controls at Llama-3.1-8B-Instruct and $B=4\mathrm{k}$. ``Macro'' averages all four benchmarks. Same-pool rows reuse exactly the same Stage-I candidates and change only the selector.}
\label{tab:targeted_controls}
\begin{tabular}{@{}lr@{}}
\toprule
Method / control & Macro (\%) \\
\midrule
Adaptive-Consistency~\citep{aggarwal2023adaptive} & 58.6 \\
TALE-EP~\citep{han2025tale} & 59.1 \\
League (fixed schedule) & 58.9 \\
\rowcolor{gray!10}\textbf{BOSS} & \textbf{60.5} \\
\midrule
No novelty transition & 59.0 \\
Fixed generation--selection split & 59.3 \\
\midrule
Same pool: plurality & 58.4 \\
Same pool: randomized A/B + repeated & 60.3 \\
\rowcolor{gray!10}Same pool: \textbf{full BOSS selector} & \textbf{60.5} \\
\bottomrule
\end{tabular}
\end{table}

The same-pool rows isolate selection from generation: changing only the selector raises macro accuracy from 58.4 with plurality to 60.3 with randomized A/B repeated voting, while the full BOSS selector reaches 60.5. Among cases where plurality is wrong but the correct hypothesis is already present, repeated judging recovers 22.8\% of errors with a 2.0\% harm rate on initially correct plurality cases; the full BOSS selector reaches 25.1\% recovery with 2.2\% harm. The fixed-split and no-novelty controls further show that, in this setting, online allocation contributes beyond a static split or a simpler switching rule. Detailed selector variants, recovery/harm analysis, and the full component ablation are reported in Appendix~\ref{sec:additional_diagnostics}. Together, these controls isolate the source of the gain: BOSS decides when residual tokens are better spent on adjudication than on another sample, while randomized repeated judging realizes most of the same-pool selection improvement.

\subsection{Hard Caps versus Realized Token Cost}
\label{sec:realized_cost}

A common hard cap equalizes the deployment constraint, not realized token use, because adaptive methods can stop early. We therefore complement the hard-cap results with approximately matched mean-token comparisons (Table~\ref{tab:realized_cost}). At the two higher-cost operating points, BOSS improves over TALE-EP by 0.3 points around 3.1k tokens and over League by 1.4 points around 3.85k; around 2.5k tokens, Adaptive-Consistency is ahead by 0.2 points. These operating points characterize a budget-dependent trade-off: BOSS is most effective when the available compute can support both candidate exploration and repeated adjudication.

\begin{table}[H]
\centering
\small
\caption{Approximately cost-matched operating points for Llama-3.1-8B.}
\label{tab:realized_cost}
\begin{tabular}{@{}lrr@{}}
\toprule
Method & Mean tokens & Macro (\%) \\
\midrule
BOSS & $\approx$2,500 & 58.4 \\
Adaptive-Consistency & 2,508 & \textbf{58.6} \\
\midrule
BOSS & $\approx$3,100 & \textbf{59.4} \\
TALE-EP & 3,074 & 59.1 \\
\midrule
BOSS & $\approx$3,850 & \textbf{60.3} \\
League & 3,853 & 58.9 \\
\bottomrule
\end{tabular}
\end{table}

Appendix~\ref{sec:additional_diagnostics} reports the full $B=4\mathrm{k}$ realized-cost audit across methods and the generation/adjudication allocation across budgets. At the common $4\mathrm{k}$ cap, BOSS uses 3,947 mean tokens and attains the highest macro accuracy in that audit (60.5), clarifying that its primary objective is productive use of the deployment allowance rather than minimum average token consumption.

\begin{figure*}[t]
  \centering
  \begin{minipage}[t]{0.228\textwidth}
    \centering
    \includegraphics[width=\linewidth,trim=285 145 2656 380,clip]{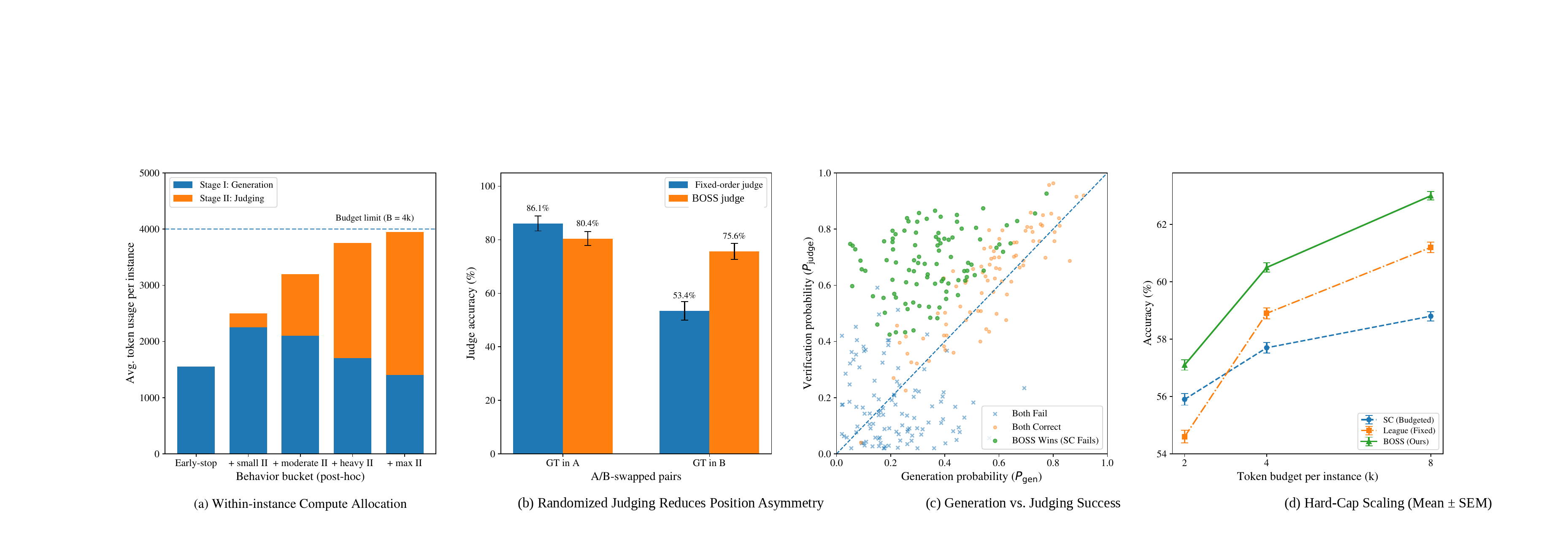}\\[-0.2ex]
    {\fontsize{7.4}{8.2}\selectfont (a) Within-instance Allocation}
  \end{minipage}\hfill
  \begin{minipage}[t]{0.230\textwidth}
    \centering
    \includegraphics[width=\linewidth,trim=1090 145 1846 380,clip]{latex/boss_mechanism_analysis_clean.png}\\[-0.2ex]
    {\fontsize{7.4}{8.2}\selectfont (b) Position Sensitivity of Judging}
  \end{minipage}\hfill
  \begin{minipage}[t]{0.230\textwidth}
    \centering
    \includegraphics[width=\linewidth,trim=1890 145 1046 380,clip]{latex/boss_mechanism_analysis_clean.png}\\[-0.2ex]
    {\fontsize{7.4}{8.2}\selectfont (c) Generation vs. Judging Success}
  \end{minipage}\hfill
  \begin{minipage}[t]{0.253\textwidth}
    \centering
    \includegraphics[width=\linewidth,trim=2690 145 166 380,clip]{latex/boss_mechanism_analysis_clean.png}\\[-0.2ex]
    {\fontsize{7.4}{8.2}\selectfont (d) Hard-Cap Scaling (Mean $\pm$ SEM)}
  \end{minipage}
  \caption{\textbf{Mechanism diagnostics for BOSS.} (a) realized generation--adjudication allocation; (b) judge-order sensitivity; (c) generation versus judging success; (d) hard-cap scaling (cap-matched).}
  \label{fig:mechanism}
\end{figure*}

\subsection{Why and When Does BOSS Switch?}
\label{sec:robustness_transition}

The transition statistics show the intended budget-adaptive shift from generation to adjudication: Stage-II entry rises from 31.8\% at $B=2\mathrm{k}$ to 84.1\% at $4\mathrm{k}$ and 94.1\% at $8\mathrm{k}$, while the safety-fallback rate falls from 65.6\% to 12.4\% and 1.2\%. Generation use grows only modestly as the cap increases, whereas most additional budget is redirected to judging; Appendix Table~\ref{tab:transition_rates} gives the full accounting.

Performance remains stable across nearby transition settings. For $\rho_{\mathrm{novel}}\in\{0.125,0.250,0.375\}$, macro accuracy is 60.1, 60.5, and 60.3, respectively; across the tested $W$ and $\tau_{\mathrm{cons}}$ ranges, it remains within 59.9--60.5. Appendix~\ref{sec:additional_diagnostics} reports the full sensitivity sweeps and token-allocation audit.

\subsection{Interpretation}
Taken together, these results support three conclusions. First, allocation matters even when the generator and judge are the same base model, so the hard-cap gains arise without a stronger external verifier. Second, adjudication contributes beyond candidate generation itself: holding the Stage-I pool fixed improves over plurality, while the fixed-split and no-novelty controls underperform the online policy. Third, adjudication has the greatest value when the available budget can support repeated comparison: BOSS leads at the two higher-cost matched operating points, while Adaptive-Consistency is ahead by 0.2 points in the lowest-cost neighborhood. BOSS therefore uses adjudication selectively when the observed candidate state and remaining budget justify its cost.

Strict accounting is central to this interpretation. Each comparison repeatedly re-encodes the problem and representative solutions, so its prompt cost is part of the allocation decision. Charging those tokens to the same ledger makes the comparison budget-faithful and ensures that the observed gains reflect how the allowance is allocated between generation and adjudication. The matched-cost results then clarify the regime in which that allocation is most useful.

\paragraph{Evidence coverage.}
Table~\ref{tab:main_results} provides the broad hard-cap comparison across both Llama-3.1 model scales and all four benchmarks, testing whether the allocation advantage persists across task, model scale, and budget. The targeted selector controls, recovery/harm analysis, approximately matched-cost operating points, and transition sensitivity are diagnostic studies centered on Llama-3.1-8B at $B=4\mathrm{k}$ unless stated otherwise. Their role is complementary: they isolate the allocation mechanism, quantify when adjudication corrects plurality errors, and characterize how the controller responds to available budget. The resulting empirical conclusion is about within-instance generation--adjudication allocation under strict per-instance caps.

\section{Conclusion}
We studied a central allocation problem in test-time reasoning: under one strict per-instance token cap, when should residual compute be spent on another candidate and when should it be spent adjudicating candidates already observed? BOSS implements a budget-safe three-way policy---return, generate, or adjudicate---using established stopping and judging primitives. Under hard caps, it is strongest in 22 of 24 model--task--budget settings in our evaluation. Same-pool controls show that the gain extends beyond candidate generation: randomized A/B judging with repeated voting raises macro accuracy from 58.4 with plurality to 60.3, and the full BOSS selector reaches 60.5. Approximately matched-cost controls further show that the value of adjudication grows with the budget available for comparison. Together, these results establish \emph{generation--adjudication allocation as a distinct and empirically useful design axis for budget-faithful test-time reasoning}. More broadly, they suggest that test-time scaling under deployment caps should treat answer selection as a first-class budgeted action, rather than a fixed post-processing stage.

\section*{Limitations}
The current evaluation covers Llama-3.1-8B-Instruct and Llama-3.1-70B-Instruct on GSM8K, MATH, GPQA-Diamond, and MMLU-Pro; extending the study to newer reasoning-specialized model families, recent mathematics benchmarks, coding, and agentic tasks is future empirical work. BOSS assumes deterministically parsable answer hypotheses, so open-ended generation would require semantic clustering or equivalence judgments before answer-level adjudication. We study \emph{within-instance} allocation under a fixed per-instance cap, complementary to distributing a global budget across questions of different difficulty~\citep{wang2025dsc}. Token count is also a deployment proxy rather than a complete serving-cost model: parallelism, prefill/decode asymmetry, cache reuse, and hardware utilization can change latency and monetary cost. Finally, the switching rule is empirically tuned rather than theoretically optimal, and successful adjudication depends on the candidate pool containing a correct hypothesis and the judge distinguishing close alternatives.

\bibliography{custom,verified_additions}

\appendix
\section{Additional Experimental Details}
This appendix provides additional implementation details for the fixed-budget evaluation, including answer parsing, prompt interfaces, budget accounting, baseline implementation, BOSS scheduling details, and hyperparameter selection.

\subsection{Answer Parsing and Normalization}
All methods use the same deterministic parser $\phi$ defined in \S\ref{sec:method}. Parsing failures are mapped to $\bot$ and counted as incorrect in all metrics. For GSM8K and MATH, we apply standard exact-match normalization to the parsed answer string, including removal of surrounding whitespace, commas, units when applicable, formatting wrappers, and common \LaTeX{} artifacts. For GPQA-Diamond and MMLU-Pro, we parse a single multiple-choice option; outputs with no valid option or ambiguous option extraction are mapped to $\bot$. These procedures are applied uniformly to BOSS and all baselines.

\subsection{Prompt Interfaces}
\paragraph{Generation prompt.} \textsc{Generate} uses a fixed instruction-following template that presents the input instance $x$ and asks the model to produce a reasoning trace followed by a final answer in a standardized format required by the parser $\phi$. The same generation template is used by all methods that perform deterministic or stochastic generation, so performance differences are attributable to inference allocation rather than prompt engineering.

\paragraph{Judge prompt.} \textsc{Judge} uses a fixed comparison template
\begin{equation}
p^{\mathrm{cmp}}(x,\mathrm{rep}(a),\mathrm{rep}(b)),
\end{equation}
which presents the input instance and two representative candidate solutions as \texttt{A} and \texttt{B}. The judge is asked to choose which candidate is more likely to contain the correct final answer. Judge decoding is deterministic with temperature $0$ and constrained to output a single token in $\{\texttt{A},\texttt{B}\}$. Any invalid output is treated as an abstention. The same judge interface is used by BOSS and all comparison-based baselines.

\subsection{Strict Budget Accounting}
All methods are evaluated under a hard per-instance token budget $B$ that counts all prompt and completion tokens. Token counts are computed with the deployed tokenizer of the corresponding base model. Each model call is executed with a maximum new-token cap. Before issuing a call with prompt $p$ and cap $L$, we compute the worst-case cost
\begin{equation}
\overline{c}(p;L)=\mathrm{tok}(p)+L.
\end{equation}
The call is issued only if
\begin{equation}
C+\overline{c}(p;L)\le B,
\end{equation}
where $C$ is the cumulative realized cost so far. Because the realized completion length never exceeds the cap, this feasibility check guarantees that the hard token budget is never violated.

\paragraph{Parallel-stream accounting.} For methods that use concurrent generations, namely FFS and short-$m@k$, token costs are accumulated across all active streams until the method terminates. Tokens consumed by streams that do not provide the returned output are still charged to the same per-instance budget.

\paragraph{Stage-transition feasibility.} BOSS enters Stage~II only when the novelty-triggered switch fires, at least two distinct parsed answer hypotheses are available, and the remaining budget can support at least one worst-case-feasible judge call. If the switch has not fired and another generation call becomes infeasible, BOSS returns the Stage-I plurality answer, or $\bot$ if no valid parsed answer exists.

\paragraph{Tie handling.} All ties are resolved deterministically to avoid hidden stochasticity beyond explicit model sampling and randomized A/B presentation. Stage-I plurality ties are broken by shorter supporting completion length and then lexicographic order. Final Stage-II ties are broken by larger Stage-I frequency, shorter representative length, and then lexicographic order.

\subsection{Baseline Implementation}
All baselines use the same token accounting, decoding caps, parsing interface, and budget-feasibility checks as BOSS.

\paragraph{CoT@1 (Greedy).} CoT@1 performs one deterministic generation with temperature $0$ using the shared generation prompt. If the hard budget is insufficient for the worst-case cost of the call, the method returns $\bot$.

\paragraph{Self-Consistency.} Self-Consistency repeatedly samples candidate completions until the next generation call is no longer budget-feasible. The final answer is selected by majority vote over parsed non-$\bot$ answers, with the same deterministic tie-breaking rule as BOSS.

\paragraph{ESC.} ESC applies an answer-convergence early-stopping rule under the same budget, generation cap, prompt template, and parsing interface. It stops once the parsed answer distribution is sufficiently stable, but does not use pairwise judging.

\paragraph{First Finish Search.} FFS launches $k$ parallel generation streams and returns the first completed output. All tokens consumed by active streams up to termination are counted toward the shared per-instance budget.

\paragraph{short-$m@k$.} short-$m@k$ launches $k$ parallel generation streams, stops when the first $m$ completions finish, and majority-votes among those $m$ outputs. Tokens consumed by all active streams before the $m$-th completion are charged to the budget.

\paragraph{Knockout and League.} Knockout and League are fixed-schedule pairwise-comparison baselines. They use the same representative-based comparison prompt, constrained judge decoding, randomized A/B ordering, repeated voting protocol, decoding caps, and token accounting as BOSS. Their generation--selection split is fixed after development tuning, whereas BOSS decides online when to transition from sampling to selection.

\subsection{Detailed BOSS Scheduling Rule}
\label{sec:detailed_scheduling}
The main text describes BOSS's stages at a high level. Algorithm~\ref{alg:boss} gives the complete budget-safe procedure; the equations below specify the Stage-II scheduling and pruning rule used in our implementation. In the pseudocode, \textsc{SwitchToSelect} combines the low-novelty transition with Stage-II feasibility, \textsc{CanMatch} applies the worst-case budget check to the current comparison prompts, and \textsc{JudgeMatch} uses randomized A/B ordering with up to $R$ repeated votes.

\begin{algorithm}[t]
\caption{\textbf{BOSS}: Budgeted Online Sampling and Selection}
\label{alg:boss}
\small
\begin{algorithmic}[1]
\Require input $x$, token budget $B$, caps $L^{\mathrm{gen}},L^{\mathrm{judge}}$
\Ensure final answer $\hat{a}$
\State $C\leftarrow 0$, $S\leftarrow \emptyset$, $q_{\mathrm{sel}}\leftarrow \mathrm{false}$

\Statex \textbf{Stage I: Adaptive candidate sampling}
\While{\Call{Feasible}{$C,p^{\mathrm{gen}}(x),L^{\mathrm{gen}},B$}}
    \State $y\leftarrow \Call{Generate}{x,L^{\mathrm{gen}}}$
    \State $C\leftarrow C+\mathrm{tok}(p^{\mathrm{gen}}(x))+\mathrm{tok}(y)$
    \State $S\leftarrow S\cup\{(y,\phi(y))\}$
    \If{\Call{Consensus}{$S$}}
        \State \Return \Call{Plurality}{$S$}
    \EndIf
    \If{\Call{SwitchToSelect}{$C,S,B$}}
        \State $q_{\mathrm{sel}}\leftarrow \mathrm{true}$; \textbf{break}
    \EndIf
\EndWhile
\If{\textbf{not} $q_{\mathrm{sel}}$}
    \State \Return \Call{PluralityOrFail}{$S$}
\EndIf

\Statex \textbf{Stage II: Budgeted pairwise selection}
\State $\mathcal{H}\leftarrow \Call{TopKAnswers}{S,K_{\max}}$
\If{$|\mathcal{H}|<2$}
    \State \Return \Call{PluralityOrFail}{$S$}
\EndIf
\For{$a\in\mathcal{H}$}
    \State $\mathrm{rep}(a)\leftarrow \Call{Representative}{a,S}$
\EndFor
\If{\textbf{not} \Call{CanMatch}{$C,\mathcal{H},\mathrm{rep},B$}}
    \State \Return \Call{PluralityOrFail}{$S$}
\EndIf

\State $\mathcal{H}_{\mathrm{act}}\leftarrow\mathcal{H}$; initialize match statistics
\While{$|\mathcal{H}_{\mathrm{act}}|\ge 2$}
    \If{\textbf{not} \Call{CanMatch}{$C,\mathcal{H}_{\mathrm{act}},\mathrm{rep},B$}}
        \State \textbf{break}
    \EndIf
    \State $(a,b)\leftarrow \Call{SelectPair}{\mathcal{H}_{\mathrm{act}}}$
    \State $(w,C)\leftarrow \Call{JudgeMatch}{a,b,\mathrm{rep},R,C,B}$
    \State Update match statistics with winner $w$
    \State $\mathcal{H}_{\mathrm{act}}\leftarrow \Call{Prune}{\mathcal{H}_{\mathrm{act}}}$
\EndWhile

\State \Return \Call{BestAnswer}{$\mathcal{H}_{\mathrm{act}},S$}
\end{algorithmic}
\end{algorithm}

After Stage~I, BOSS constructs the hypothesis set
\begin{equation}
\mathcal{H}_x=\{a\in\mathcal{A}:n_a>0\}
\end{equation}
and retains the $K_{\max}$ most frequent parsed answers:
\begin{equation}
\mathcal{H}\leftarrow\mathrm{TopK}(\mathcal{H}_x;K_{\max}).
\end{equation}
For each retained answer $a\in\mathcal{H}$, BOSS selects the shortest supporting completion with at least $L^{\mathrm{rep}}_{\min}$ tokens; if none qualifies, it uses the shortest supporting completion. Representatives longer than $L^{\mathrm{rep}}$ tokens are truncated while preserving the final-answer line.

For each active hypothesis $a$, BOSS maintains a win count $w_a$ and a match count $g_a$, initialized to zero. After a match between $a$ and $b$, both match counts increase by one. The winner receives one win; if the match is tied, each answer receives half a win. BOSS computes the smoothed win estimate
\begin{equation}
\hat{p}_a=\frac{w_a+\alpha}{g_a+2\alpha},
\end{equation}
and the lightweight uncertainty radius
\begin{equation}
r_a=\sqrt{\frac{\log(2|\mathcal{H}|/\delta)}{2(g_a+1)}}.
\end{equation}
It then defines
\begin{align}
\mathrm{UCB}_a&=\min(1,\hat{p}_a+r_a),\\
\mathrm{LCB}_a&=\max(0,\hat{p}_a-r_a).
\end{align}
These quantities are used only as scheduling and pruning heuristics, not as calibrated confidence bounds.

Let $\mathcal{H}_{\mathrm{act}}$ be the active hypothesis set. While at least two active hypotheses remain and another judge call is budget-feasible, BOSS selects the two distinct active hypotheses with the largest UCB values and compares them using randomized A/B ordering with repeated voting. After updating match statistics, BOSS prunes clearly dominated hypotheses:
\begin{equation}
\ell^\star=\max_{u\in\mathcal{H}_{\mathrm{act}}}\mathrm{LCB}_u,
\end{equation}
\begin{equation}
\mathcal{H}_{\mathrm{act}}\leftarrow
\mathcal{H}_{\mathrm{act}}\setminus\{u:\mathrm{UCB}_u<\ell^\star\}.
\end{equation}
If pruning would remove all active hypotheses, BOSS retains the answer with the largest pre-pruning UCB. When no further budget-feasible comparison is possible, BOSS returns the active hypothesis with the largest $\hat{p}_a$, using the deterministic tie-breaking rule described above.

\subsection{Hyperparameter Selection}
All allocation-sensitive hyperparameters are selected on held-out development subsets that are disjoint from the reported evaluation data. This protocol is applied uniformly to BOSS and all tunable baselines. It covers the parallelism parameters $k$ and $m$ for FFS and short-$m@k$, the fixed generation--selection splits for Knockout and League, and the stage-transition and selection parameters of BOSS. No reported test instances are used for hyperparameter selection.

Table~\ref{tab:hyperparams} lists the BOSS hyperparameters used in our experiments. The generation and judge decoding caps are fixed across benchmarks, while the allocation thresholds are selected on development data and then fixed before test evaluation.

\begin{table}[t]
\centering
\small
\caption{BOSS hyperparameters. Values are fixed after development tuning and are not selected using reported test instances.}
\label{tab:hyperparams}
\begin{tabular}{@{}lc@{}}
\toprule
Hyperparameter & Value \\
\midrule
Generation cap $L^{\mathrm{gen}}$ & 1024 \\
Judge cap $L^{\mathrm{judge}}$ & 4 \\
Judge replicates $R$ & 3 \\
Consensus threshold $\tau_{\mathrm{cons}}$ & 0.70 \\
Consensus gap $m_{\mathrm{gap}}$ & 2 \\
Novelty window $W$ & 4 \\
Novelty threshold $\rho_{\mathrm{novel}}$ & 0.25 \\
Maximum retained hypotheses $K_{\max}$ & 5 \\
Representative minimum length $L^{\mathrm{rep}}_{\min}$ & 32 \\
Representative cap $L^{\mathrm{rep}}$ & 256 \\
Smoothing constant $\alpha$ & 1.0 \\
Scheduling parameter $\delta$ & 0.05 \\
\bottomrule
\end{tabular}
\end{table}

\subsection{Random Seeds and Reporting}
All stochastic methods are evaluated with 5 random seeds. Tables report mean accuracy over seeds unless otherwise stated. When error bars are shown in figures, they denote mean $\pm$ standard error of the mean (SEM). Deterministic components, including parsing, tie-breaking, budget-feasibility checks, and constrained judge decoding, are kept fixed across methods.

\subsection{Scope of the Evaluation}
Our evaluation focuses on reasoning benchmarks with deterministically parsable final answers and a fixed per-instance deployment cap. This scope matches the design of BOSS, which groups candidate completions by parsed answer hypotheses and performs answer-level adjudication. Cross-instance budget allocation, semantic clustering for open-ended outputs, and evaluation on newer reasoning-specialized model families are outside the current evidence.

\section{Additional Diagnostics}
\label{sec:additional_diagnostics}

The following analyses provide supporting evidence for the main-text controls without changing the central claim.

\subsection{Statistical Uncertainty at the Closest Setting}
\begin{table}[H]
\centering
\small
\caption{Llama-3.1-8B-Instruct at $B=4\mathrm{k}$: mean accuracy (\%) with 95\% confidence intervals, plus paired per-instance $p$-values against the strongest non-BOSS baseline for each benchmark.}
\label{tab:uncertainty}
\begin{adjustbox}{width=\columnwidth}
\begin{tabular}{@{}llrrr@{}}
\toprule
Benchmark & Baseline & Baseline (95\% CI) & BOSS (95\% CI) & Paired $p$ \\
\midrule
GSM8K & short-$m@k$ & 87.9 [87.5, 88.3] & 88.8 [88.4, 89.2] & 0.021 \\
MATH & League & 57.2 [56.7, 57.7] & 59.4 [58.8, 60.0] & $<0.001$ \\
GPQA-D & League & 37.8 [36.9, 38.7] & 39.5 [38.5, 40.5] & 0.026 \\
MMLU-Pro & League & 52.9 [52.5, 53.3] & 54.4 [53.9, 54.9] & $<0.001$ \\
\bottomrule
\end{tabular}
\end{adjustbox}
\end{table}

The intervals quantify uncertainty in the method-level accuracy estimates, while the $p$-values use paired per-instance predictions. We report paired tests for the closest Llama-3.1-8B, $B=4\mathrm{k}$ comparisons, where the margins to the strongest baselines are smallest.

\subsection{All-Method Realized-Cost Audit at B=4k}
\begin{table}[H]
\centering
\small
\caption{Llama-3.1-8B-Instruct at the common $B=4\mathrm{k}$ cap. Token statistics are realized per-instance totals; Macro averages the four benchmarks.}
\label{tab:all_method_cost}
\begin{adjustbox}{width=\columnwidth}
\begin{tabular}{@{}lrrrr@{}}
\toprule
Method & Mean & Median & P90 & Macro (\%) \\
\midrule
CoT@1 & 846 & 779 & 1330 & 53.8 \\
FFS & 2168 & 2044 & 3312 & 56.6 \\
Adaptive-Consistency & 2508 & 2365 & 3868 & 58.6 \\
ESC & 2736 & 2584 & 3924 & 58.0 \\
TALE-EP & 3074 & 3012 & 3968 & 59.1 \\
short-$m@k$ & 3246 & 3330 & 3984 & 58.2 \\
SC & 3721 & 3862 & 3995 & 57.7 \\
Knockout & 3809 & 3894 & 3996 & 58.2 \\
League & 3853 & 3926 & 3997 & 58.9 \\
\rowcolor{gray!10}\textbf{BOSS} & \textbf{3947} & \textbf{3981} & \textbf{3999} & \textbf{60.5} \\
\bottomrule
\end{tabular}
\end{adjustbox}
\end{table}

This audit makes the cost trade-off explicit: BOSS uses a larger share of the common $4\mathrm{k}$ allowance, consistent with its objective of reallocating residual budget rather than minimizing mean token use. The primary hard-cap comparison therefore measures how effectively methods use a shared deployment allowance, while Table~\ref{tab:realized_cost} provides the complementary approximately matched-cost view.

\begin{table*}[t]
\centering
\small
\caption{Mean realized token utilization over 5 random seeds. All prompt and completion tokens are counted.}
\label{tab:utilization}
\begin{tabular}{@{}cccccccccccc@{}}
\toprule
& \multicolumn{5}{c}{\textbf{League} (Fixed)} && \multicolumn{5}{c}{\textbf{BOSS} (Adaptive)} \\
\cmidrule(lr){2-6}\cmidrule(lr){8-12}
$B$ & Gen. & Judge & Total & Util.\% & Judge\% && Gen. & Judge & Total & Util.\% & Judge\% \\
\midrule
$1\mathrm{k}$ & 523 & 414 & 937 & 93.7 & 44.2 && 958 & 0 & 958 & 95.8 & 0.0 \\
$2\mathrm{k}$ & 812 & 1103 & 1915 & 95.8 & 57.6 && 1445 & 492 & 1937 & 96.9 & 25.4 \\
$4\mathrm{k}$ & 1634 & 2219 & 3853 & 96.3 & 57.6 && 1903 & 2044 & 3947 & 98.7 & 51.8 \\
$8\mathrm{k}$ & 2542 & 5237 & 7779 & 97.2 & 67.3 && 1987 & 5924 & 7911 & 98.9 & 74.9 \\
\bottomrule
\end{tabular}
\end{table*}

\subsection{Detailed Same-Pool Selection Analysis}
\begin{table}[H]
\centering
\small
\caption{Detailed targeted controls at Llama-3.1-8B-Instruct, $B=4\mathrm{k}$. ``Macro'' averages all four benchmarks; same-pool rows reuse identical Stage-I candidates and change only the selector.}
\label{tab:targeted_controls_full}
\begin{adjustbox}{width=\columnwidth}
\begin{tabular}{@{}lr@{}}
\toprule
Method / control & Macro (\%) \\
\midrule
\multicolumn{2}{l}{\textit{Competing adaptive policies}} \\
Adaptive-Consistency~\citep{aggarwal2023adaptive} & 58.6 \\
TALE-EP~\citep{han2025tale} & 59.1 \\
League (fixed schedule) & 58.9 \\
\rowcolor{gray!10}\textbf{BOSS} & \textbf{60.5} \\
\midrule
\multicolumn{2}{l}{\textit{Allocation controls}} \\
No novelty transition & 59.0 \\
Fixed generation--selection split & 59.3 \\
\midrule
\multicolumn{2}{l}{\textit{Same Stage-I candidate pool}} \\
Plurality & 58.4 \\
Single fixed-order judge & 59.1 \\
Randomized A/B judge & 59.7 \\
Randomized A/B + repeated voting & 60.3 \\
\rowcolor{gray!10}\textbf{Full BOSS selector} & \textbf{60.5} \\
\bottomrule
\end{tabular}
\end{adjustbox}
\end{table}

\begin{table}[H]
\centering
\small
\caption{Error-correction behavior on identical Stage-I candidate pools. Recovery is measured when plurality is wrong but the correct hypothesis is present; harm is measured when plurality is initially correct.}
\label{tab:recovery_harm}
\begin{tabular}{@{}lrr@{}}
\toprule
Selector & Recovery (\%) & Harm (\%) \\
\midrule
Randomized A/B + repeated & 22.8 & 2.0 \\
Full BOSS selector & \textbf{25.1} & 2.2 \\
\bottomrule
\end{tabular}
\end{table}

\subsection{Full Component Ablation}
\begin{table}[H]
\centering
\small
\caption{Component ablation at $B=4\mathrm{k}$ with Llama-3.1-8B-Instruct. Selected columns are shown; ``Avg'' averages all four benchmarks.}
\label{tab:ablation_full}
\begin{tabular}{@{}lccc@{}}
\toprule
Variant & MATH & GPQA-D & Avg \\
\midrule
\rowcolor{gray!10}\textbf{BOSS (Full)} & \textbf{59.4} & \textbf{39.5} & \textbf{60.5} \\
No early stop & 58.0 & 38.7 & 59.6 \\
No novelty transition & 58.1 & 37.6 & 59.0 \\
Fixed gen.--sel. split & 58.5 & 38.1 & 59.3 \\
Fixed-order single-vote judge & 56.8 & 35.0 & 57.7 \\
Round-robin scheduling & 57.4 & 37.1 & 58.8 \\
Full representatives & 55.9 & 38.3 & 58.3 \\
\bottomrule
\end{tabular}
\end{table}

\subsection{Mechanism and Allocation Diagnostics}

\begin{table}[H]
\centering
\small
\caption{Observed BOSS allocation by hard budget. Gen./Judge are mean realized tokens; Entry/Fallback are trajectory rates.}
\label{tab:transition_rates}
\begin{tabular}{@{}lrrrr@{}}
\toprule
$B$ & Gen. & Judge & Entry (\%) & Fallback (\%) \\
\midrule
$2\mathrm{k}$ & 1445 & 492 & 31.8 & 65.6 \\
$4\mathrm{k}$ & 1903 & 2044 & 84.1 & 12.4 \\
$8\mathrm{k}$ & 1987 & 5924 & 94.1 & 1.2 \\
\bottomrule
\end{tabular}
\end{table}

\begin{table}[H]
\centering
\scriptsize
\caption{Sensitivity of Llama-3.1-8B-Instruct at $B=4\mathrm{k}$ to the principal transition hyperparameters. Bold marks the main configuration. Entry is the Stage-II entry rate; for $\tau_{\mathrm{cons}}$, parentheses also give the consensus-exit rate.}
\label{tab:sensitivity_full}
\begin{adjustbox}{width=\columnwidth}
\begin{tabular}{@{}llrr@{}}
\toprule
Parameter & Value & Macro (\%) & Behavior (\%) \\
\midrule
$\rho_{\mathrm{novel}}$ & 0.000 & 59.4 & entry 52.6 \\
 & 0.125 & 60.1 & entry 72.8 \\
 & \textbf{0.250} & \textbf{60.5} & entry 84.1 \\
 & 0.375 & 60.3 & entry 89.7 \\
 & 0.500 & 59.8 & entry 93.2 \\
\midrule
$W$ & 2 & 60.0 & entry 90.5 \\
 & 3 & 60.4 & entry 86.9 \\
 & \textbf{4} & \textbf{60.5} & entry 84.1 \\
 & 6 & 60.2 & entry 77.8 \\
 & 8 & 59.9 & entry 70.4 \\
\midrule
$\tau_{\mathrm{cons}}$ & 0.60 & 59.9 & entry 79.0 (cons. 8.9) \\
 & \textbf{0.70} & \textbf{60.5} & entry 84.1 (cons. 3.5) \\
 & 0.80 & 60.4 & entry 85.7 (cons. 1.8) \\
 & 0.90 & 60.0 & entry 86.6 (cons. 0.7) \\
\bottomrule
\end{tabular}
\end{adjustbox}
\end{table}

Across the tested switching hyperparameters, macro-average accuracy remains within 59.8--60.5 for the full $\rho_{\mathrm{novel}}$ sweep and within 59.9--60.5 across the tested $W$ and $\tau_{\mathrm{cons}}$ values. The behavior columns show the expected directional effects on switching and early consensus.

\section*{Use of Large Language Models}
LLMs assisted with language polishing, presentation, and manuscript organization; the authors verified all experiments, analyses, results, and scientific claims.

\end{document}